\documentclass[letterpaper,10pt,conference]{ieeeconf}
\IEEEoverridecommandlockouts
\usepackage{amsmath,amssymb,graphicx,booktabs,array,multirow,xcolor,url}
\usepackage{tabularx}
\newcommand{\method}{ConsensusTAS}

\title{\LARGE\bf ConsensusTAS: Self-Supervised Temporal Action Segmentation for Long-Horizon Construction Videos
}

\author{Xiaoshan Zhou$^{1}$ and Yafei Sun$^{2}$%
\thanks{Corresponding author: Xiaoshan Zhou}%
\thanks{$^{1}$Xiaoshan Zhou is with the School of Project Management, University of Sydney, Australia
{\tt\small xiaoshan.zhou@sydney.edu.au}}%
\thanks{$^{2}$Yafei Sun is with the School of Civil and Environmental Engineering, University of New South Wales, Australia
{\tt\small yafei.sun@unsw.edu.au}}}

\begin{document}
\flushbottom
\maketitle
\thispagestyle{empty}
\pagestyle{empty}

\begin{abstract}
Recognizing sequential construction activities is important for collaborative human–robot work; for example, robots are able to understand workers’ current and upcoming actions and provide timely tool delivery or physical support. However, despite extensive research on construction worker activity recognition, existing studies have been limited to classifying activity categories, such as climbing, lifting, and walking, instead of recognizing fine-grained activity transitions from long-horizon sequences. Addressing this problem is challenging because annotating action temporal boundaries in long construction videos is time-consuming. In this study, we propose ConsensusTAS, a label-free, self-supervised learning approach to segment continuous video streams into distinct activity phases by exploiting the internal consensus of candidate segmentations. We evaluated our algorithm on three public datasets, where it outperformed state-of-the-art methods, achieving an F1@10 of 73.08 on GTEA, an F1@10 of 64.33 on Breakfast, and an F1@50 of 33.50 on static-camera videos from Assembly101. We also tested it on real-world construction videos, where post-hoc evaluation showed that the model successfully recognized and segmented actions within the composite activity of bricklaying, such as spreading mortar on a brick, placing the brick, pressing, and aligning. Compared with other temporal action segmentation models that require computationally intensive large vision–language models, our method can run on a CPU, which provides practical value for video surveillance and human–robot collaboration on mobile robotic platforms.
\end{abstract}

\section{Introduction}

Anticipating workers’ next steps is essential for advancing collaborative construction robotics from supporting structured actions, such as passing a tool, toward coordinating assistance throughout an entire workflow. Current research has sought to understand which activities workers are performing, such as tiling and masonry~\cite{monfared2026onsite}. The next step is to identify their fine-grained constituent actions. For example, hand-sawing a panel involves positioning the panel, measuring and marking the cutting line, grasping the saw, and cutting. Recognizing these individual actions and their transitions offers a more detailed understanding of the workflow than classifying the entire sequence as panel cutting.
 
Temporal action segmentation (TAS) of long-horizon videos is challenging because manually annotating frame-level action labels is time-consuming~\cite{wang2022sscap}. Although several public datasets capturing activities such as breakfast preparation and tool assembly provide testbeds for TAS model development and evaluation~\cite{kuehne2014breakfast,sener2022assembly101}, no comparable dataset is currently available for the construction domain. Developing such a dataset requires videos of workers performing natural, multi-step activities on construction sites, such as bricklaying and rebar typing, and a label-free TAS method that can partition these videos into distinct activity phases, which is the purpose of methodological development in this study.

Self-supervised learning offer a promising means of reducing this annotation burden. These methods exploit information inherent in the videos, such as visual similarity between frames and temporal continuity at action boundaries, to discover recurring action phases. For example, SSCAP learns discriminative frame representations through self-supervision and then uses the co-occurrence patterns of sub-actions to infer their temporal structure~\cite{wang2022sscap}. More recent approaches introduced object-centric features to identify boundaries of changes in human–object interactions~\cite{li2024otas}, and used temporally consistent unbalanced optimal transport to generate pseudo-labels for sub-actions of varying durations~\cite{xu2024asot}.

Despite this progress, existing unsupervised TAS methods have several limitations. SSCAP learns co-occurrence relationships across a corpus of videos, making its segmentation dependent on recurring sub-action patterns~\cite{wang2022sscap}. OTAS focuses primarily on detecting boundaries from global and object-centric visual changes; however, visually salient changes do not always correspond to meaningful activity transitions, particularly in construction videos containing camera motion, occlusion, or changes in the surrounding environment~\cite{li2024otas}. Optimal-transport-based methods improve temporal consistency but require iterative optimization and may remain sensitive to the initial frame–action affinity matrix and the assumed number of latent actions~\cite{xu2024asot}. 

To address these limitations, we build on the internal-objective design of a progression-aware segmentation framework recently introduced for temporal phase discovery in time series~\cite{zhou2026ssel} and extend it to the video domain. In \method{}, candidates generated under varied temporal scales and randomized settings are scored using representation-derived internal criteria, and elite candidates are converted into a continuous boundary density from which the final segmentation is decoded. This consensus-based design reduces dependence on any single initialization or temporal scale while retaining a fully label-free workflow.

\section{Method}
\subsection{Problem Formulation}
Given a video representd by a sequence of frame-level features
\begin{equation}
X=(x_1,\ldots,x_T),\qquad x_t\in\mathbb{R}^{D},
\label{eq:input}
\end{equation}
the objective is to infer an ordered boundary set
\begin{equation}
B=\{b_1,\ldots,b_K\},\qquad 1<b_1<\cdots<b_K<T,
\label{eq:boundaries}
\end{equation}
without action labels during prediction or calibration.

Here $T$ is sequence length, $D$ is feature dimension, and $K$ is the number of boundaries. The inferred boundaries partition the video into $K+1$ anonymous temporal segments.

\subsection{Feature Preprocessing}
Each dimension is standardized using its temporal median and median absolute
deviation (MAD):
\begin{equation}
\widetilde{x}_{t,d}=\operatorname{clip}\!\left(
\frac{x_{t,d}-\operatorname{median}_{t}(x_{t,d})}
{1.4826\operatorname{MAD}_{t}(x_{t,d})+\epsilon},-8,8\right).
\label{eq:robust}
\end{equation}
Features are centered, projected to at most 32 dimensions using randomized PCA. Each projected frame vector is then L2 normalized.

\subsection{Multiscale boundary evidence}

For each sampled scale $s$ and smoothing setting, left- and right-context
representations are compared around every frame. The raw boundary score
combines cosine dissimilarity and Euclidean motion:
\begin{equation}
e_t
=
0.65\left(1-\cos(\ell_t,r_t)\right)
+
0.35\left\lVert \ell_t-r_t \right\rVert_2 .
\end{equation}

After robust normalization, the default foreground weighting is
\begin{equation}
e'_t
=
e_t
\left(
0.25
+
0.75\,\operatorname{MA}_5(\hat{f})_t
\right).
\end{equation}

\subsection{Candidate Population and Internal Reward}
The default population contains 48 candidates. Each samples a temporal scale,
smoothing width, 55--100\% of projected dimensions, and a score quantile
uniformly between 0.72 and 0.92. Gaussian perturbation has standard deviation
0.015. For candidate $i$, agreement is the mean tolerance-based boundary F1 against the remaining population:
\begin{equation}
A_i=\frac{1}{N-1}\sum_{j\neq i}F_{\rm match}(B_i,B_j).
\label{eq:agreement}
\end{equation}
Boundary contrast $C_i$ is the mean boundary-evidence at its proposed boundaries; compactness $S_i$ compares within-segment feature variance with global variance. Complexity is
\begin{equation}
P_i=\frac{|B_i|\log T}{T}.
\end{equation}
The implemented reward is
\begin{equation}
R_i=0.50A_i+0.25C_i+0.25S_i-0.08P_i.
\label{eq:reward}
\end{equation}
The top 25\% candidates are retained, with a minimum of three elites.

\subsection{Consensus Density}
Elite boundaries contribute Gaussian kernels:
\begin{equation}
d(t)=\sum_{i\in\mathcal E}w_i\sum_{b\in B_i}
\exp\!\left[-\frac{(t-b)^2}{2h^2}\right],
\label{eq:density}
\end{equation}
where
\begin{equation}
w_i=\frac{\exp((R_i-R_{\max})/0.1)}
{\sum_{j\in\mathcal E}\exp((R_j-R_{\max})/0.1)}.
\label{eq:weights}
\end{equation}
The density is normalized by its maximum. Peaks above a dataset/configuration-specific density quantile are retained while enforcing a minimum gap.

\begin{figure}[!t]
\centering
\includegraphics[width=0.735\columnwidth,height=0.165\textheight,keepaspectratio]{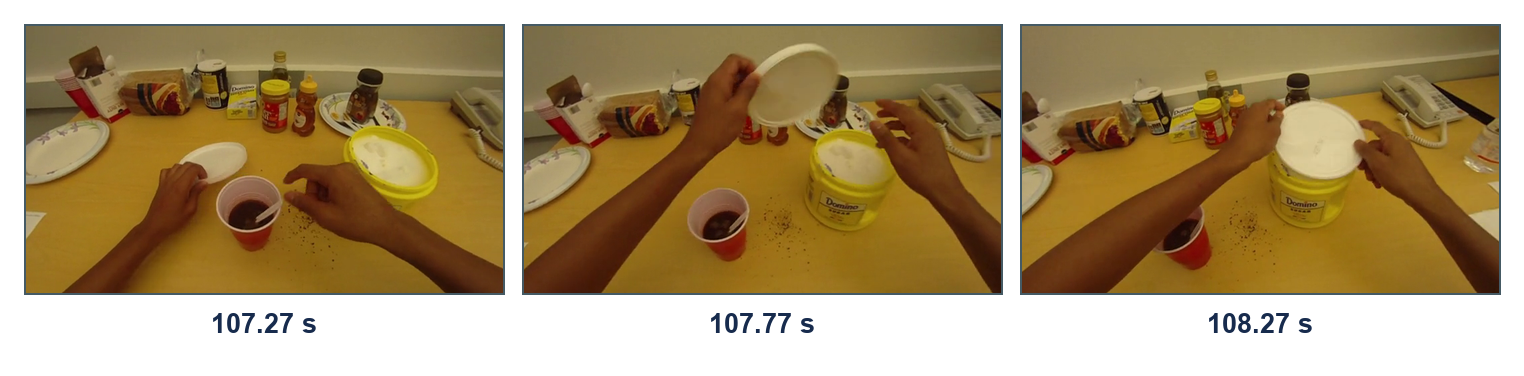}
\caption{Sample images from the GETA dataset.}
\label{fig:demo0}
\vspace{1mm}
\includegraphics[width=0.70\columnwidth,height=0.105\textheight,keepaspectratio]{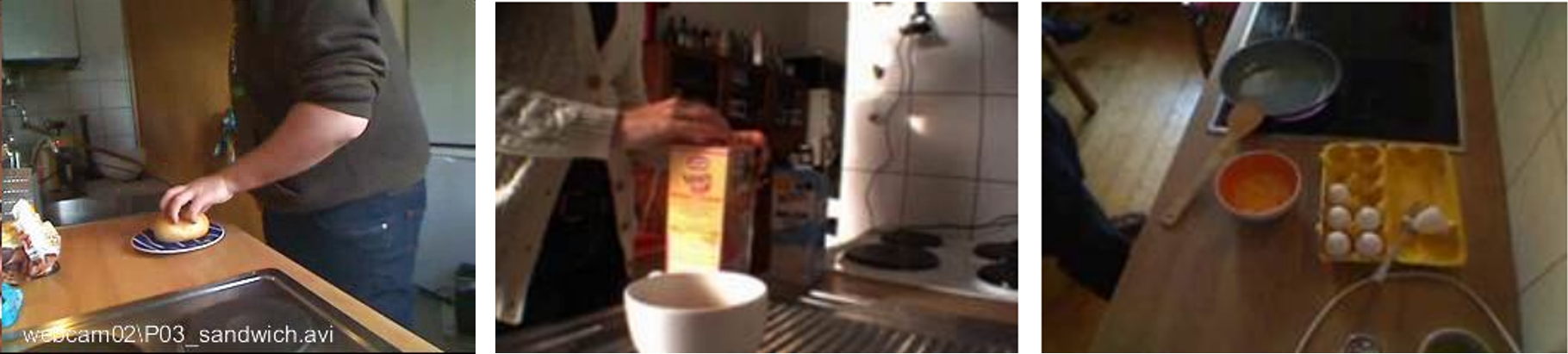}
\caption{Sample images from the Breakfast dataset.}
\label{fig:demo2}
\vspace{1mm}
\includegraphics[width=0.70\columnwidth,height=0.105\textheight,keepaspectratio]{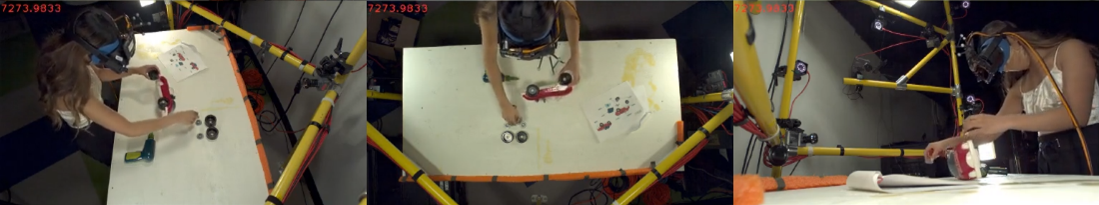}
\caption{Sample images from the Assembly101 dataset.}
\label{fig:demo3}
\end{figure}

\section{Experimental Protocol}
\subsection{Dataset}
\subsubsection{GTEA}
Georgia Tech Egocentric Activities (GTEA) contains first-person kitchen videos recorded with a head-mounted camera. Four participants each perform seven activities, including making a sandwich, tea or coffee, giving 28 videos in the
TAS version. All 28 local feature sequences and four official
splits were used. See Fig.~\ref{fig:demo0}~\cite{fathi2011gtea}.

\subsubsection{Breakfast}
The Breakfast Actions Dataset (Breakfast) records people preparing ten breakfast-related items, such as cereal, fried egg, and packages.. It contains 1712 videos from 52 participants. Breakfast is one of the most established benchmarks for TAS models and it contains more variation than GTEA in people, environments, viewpoints and action sequences. See Fig.~\ref{fig:demo2}~\cite{kuehne2014breakfast}.

\subsubsection{Assembly101}
Assembly101 is a much larger dataset involving participants assembling and disassembling 101 types of toy vehicles. In this study, we used a subset C10119, the static-view feature package with coarse action labels, such as attaching or detaching a vehicle component. It contains 350 useable recording (12 of the 362 recordings in the downloaded package lacked usable labels and were excluded) and 680 resulting assembly/disassembly activity crops. See Fig.~\ref{fig:demo3}~\cite{sener2022assembly101}.

\subsection{Evaluation Metrics}
Class-agnostic segment F1 is reported at temporal IoU thresholds
$\tau\in\{0.10,0.25,0.50\}$. Background is excluded and anonymous predicted
segments are greedily matched to reference segments:
\begin{equation}
F1@\tau=100\frac{2TP_\tau}{2TP_\tau+FP_\tau+FN_\tau}.
\label{eq:f1}
\end{equation}

\vspace{2mm}

\section{Results}

Tables~\ref{tab:gtea}, \ref{tab:breakfast}, and
\ref{tab:assembly} summarize the results on GTEA, Breakfast, and
Assembly101, respectively. Under the class-agnostic segment-overlap protocol,
ConsensusTAS achieved the strongest reported performance among the compared
unsupervised methods on GTEA and Breakfast. On GTEA, ConsensusTAS v1.1 obtained
F1 scores of 73.08, 71.07, and 52.40 at overlap thresholds of 10\%, 25\%, and
50\%, respectively. On Breakfast, it achieved 64.33, 62.45, and 44.73. The
improvement at F1@50 is particularly notable, indicating that the discovered
segments maintain greater temporal overlap with the reference phases rather
than merely recovering their approximate locations.

On the C10119 static-view subset of Assembly101, ConsensusTAS outperformed
uniform segmentation, change-point detection, KMeans with smoothing, and the
implemented ASOT variants. Its comparison with ASESM was threshold-dependent.
Relative to the retrospectively best ASESM grid configuration, ConsensusTAS had
a paired difference of $-5.01$ at F1@10, with a 95\% confidence interval of
$[-6.41,-3.58]$, but a difference of $+7.28$ at F1@50, with a 95\% confidence
interval of $[5.22,9.30]$. Thus, ASESM recovered more segments under the lenient
overlap criterion, whereas ConsensusTAS produced substantially better
high-overlap localization. The strongest ASESM configuration was
identified through an extensive parameter grid evaluated on the test set though
(See Table~\ref{tab:assemblyASESM}).

\begin{table}[!t]
\caption{GTEA results over three seeds and four folds. Baseline results are reported by Aouaidjia et
al.~\cite{aouaidjia2026similarity}.}
\label{tab:gtea}
\centering\footnotesize
\setlength{\tabcolsep}{2pt}
\renewcommand{\arraystretch}{0.90}
\begin{tabular}{@{}lc@{}}
\toprule Method & F1@10 \\
\midrule
\method\ & $\mathbf{73.08\pm2.18}$ \\ASESM, manually selected interval & 70.4\\ASESM, automatic interval & 57.0  \\SemiTAS comparison, 5\% labels & 59.8 \\SemiTAS comparison, 10\% labels & 71.5\\
\bottomrule
\end{tabular}
\end{table}

\begin{table}[!t]
\caption{Breakfast results over seeds and official folds.}
\label{tab:breakfast}
\centering\footnotesize
\setlength{\tabcolsep}{2pt}
\renewcommand{\arraystretch}{0.90}
\begin{tabular}{@{}lccc@{}}
\toprule Method & F1@10 & F1@25 & F1@50 \\
\midrule
TSA, published & 58.0 & -- & -- \\
SaM, published & 55.9 & -- & -- \\
Uniform 300, local & 63.09 & 58.41 & 33.60 \\
Uniform 350, local & 63.36 & 59.18 & 33.38 \\
ASESM, local & 63.12 & 58.59 & 32.77 \\
ASESM, published & 63.2 & -- & -- \\
\method\ & $\mathbf{64.33\pm0.70}$ & $\mathbf{62.45\pm0.95}$ & $\mathbf{44.73\pm1.31}$ \\
\bottomrule
\end{tabular}
\end{table}

\begin{table}[!t]
\caption{Assembly101 test results on 167 eligible activity crops.}
\label{tab:assembly}
\centering\footnotesize
\setlength{\tabcolsep}{1.5pt}
\renewcommand{\arraystretch}{0.90}
\begin{tabular}{@{}lccc@{}}
\toprule Method & F1@10 & F1@25 & F1@50 \\
\midrule
Uniform & 48.94 & 40.06 & 18.05 \\
Change point & 47.69 & 43.81 & 23.78  \\
KMeans + smoothing & $40.62\pm0.04$ & $37.01\pm0.06$ & $20.41\pm0.27$ \\
Adapted ASOT & $42.56\pm0.37$ & $39.50\pm0.39$ & $22.25\pm0.37$ \\
Local full ASOT & $28.84\pm4.32$ & $23.60\pm4.92$ & $11.90\pm2.88$ \\
\method & $54.79\pm0.10$ & $52.60\pm0.08$ & $\mathbf{33.50\pm0.24}$ \\
ASESM, $K=16$, gap 225 & $\mathbf{59.81\pm0.05}$ & $\mathbf{53.55\pm0.07}$ & $26.21\pm0.18$ \\
\bottomrule
\end{tabular}
\end{table}

\begin{table}[t]
\caption{Assembly101 test results on 167 eligible activity crops.}
\label{tab:assemblyASESM}
\centering\footnotesize
\setlength{\tabcolsep}{2pt}
\renewcommand{\arraystretch}{0.90}
\begin{tabular}{@{}lccc@{}}
\toprule
ASESM configuration & F1@10 & F1@25 & F1@50 \\
\midrule
Label-free selected: $K=16$, gap 30
    & 11.40 & 4.55 & 0.97 \\
Grid sensitivity: $K=16$, gap 225
    & $\mathbf{59.81}$ & $\mathbf{53.55}$ & 26.21 \\
ConsensusTAS
    & 54.79 & 52.60 & $\mathbf{33.50}$ \\
\bottomrule
\end{tabular}
\end{table}

In addition, we conducted one-component-at-a-time ablations using the Assembly101 test subset. All ablations retained the same temporal configuration: scales $(10,20,40,80)$, $L_{\min}=30$, $K_{\max}=40$,
$h=12$, $Q_{\rm density}=0.80$, and $G_{\min}=45$. Experiments used seeds 7, 17, and 27 over 167 eligible activity crops. As shown
in Table~\ref{tab:ablations}, removing density decoding reduced F1@50 from
33.50 to 28.42, while removing internal calibration reduced it
to 29.98. These results identify consensus-density decoding and temporal-scale
calibration as the most influential components on the Assembly101 dataset, while not yet excluding other heuristics' role in long, untrimmed recordings containing distinct motion dynamics or irrelevant inactivity.

\begin{table}[!t]
\caption{Assembly101 ablations. Test-set analyses are diagnostic.}
\label{tab:ablations}
\centering\footnotesize
\setlength{\tabcolsep}{1.5pt}
\renewcommand{\arraystretch}{0.90}
\begin{tabularx}{\columnwidth}{@{}>{\raggedright\arraybackslash}p{0.25\columnwidth}ccc>{\raggedright\arraybackslash}X@{}}
\toprule Variant & F1@10 & F1@25 & F1@50 & Interpretation \\
\midrule
Full \method & 54.79 & 52.60 & 33.50 & Reference \\
No consistency reward & 54.80 & 52.60 & 32.96 & Small strict-overlap loss \\
No foreground suppression & 55.49 & 53.42 & 34.38 & Foreground term unfavorable \\
Population 12 & 55.00 & 52.83 & 33.18 & Large population not essential \\
Unweighted density & 54.76 & 52.59 & 33.52 & Weighting has little effect \\
No density decoder & 48.27 & 45.73 & 28.42 & Large degradation \\
No internal calibration & 49.73 & 47.48 & 29.98 & Large degradation \\
\bottomrule
\end{tabularx}
\end{table}

\begin{figure}[!t]
\centering
\includegraphics[width=0.92\columnwidth,height=0.17\textheight,keepaspectratio]{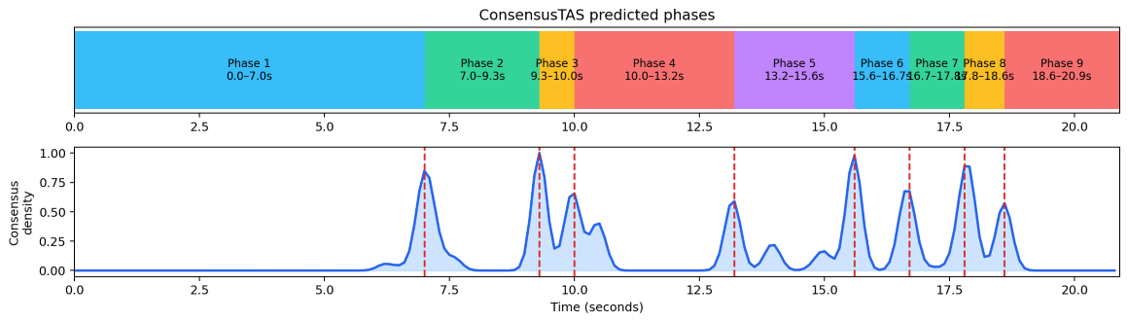}
\caption{Timeline and consensus density for the bricklaying video.}
\label{fig:demo}
\end{figure}

\begin{figure*}[!t]
    \centering
    \includegraphics[width=0.92\textwidth,height=0.17\textheight,keepaspectratio]{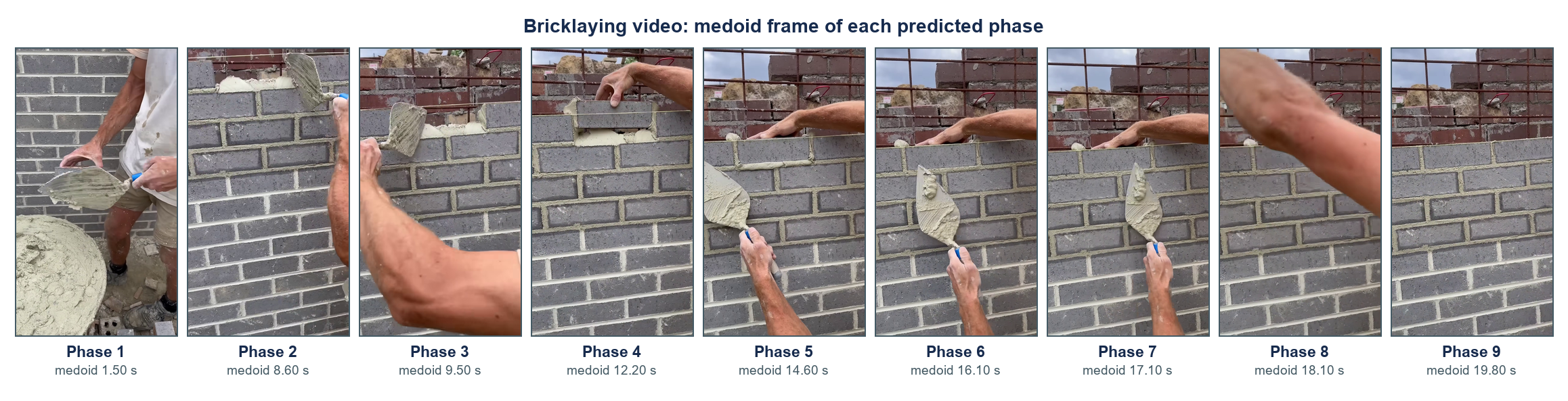}
    \caption{Qualitative visualization of the nine anonymous phases discovered
    by ConsensusTAS in an unlabeled bricklaying video.}
    \label{fig:bricklaying-medoids}
\end{figure*}

\section{Real-World Demonstrations}
We further examined whether ConsensusTAS could produce interpretable phase
boundaries in a previously unseen, unlabeled bricklaying video. This
demonstration used a training-free OpenCV front end. Frames were sampled at target 10 frames/s. Each frame was resized to $160\times160$. A normalized $16\times8$ HSV
histogram gives $h_t\in\mathbb{R}^{128}$. Grayscale structure and Canny edges
(thresholds 60 and 140) are resized to $16\times16$, giving
$g_t,e_t\in\mathbb{R}^{256}$. Farneback flow uses pyramid scale 0.5, three
levels, window 15, three iterations, polynomial neighborhood 5, and sigma 1.2. Flow magnitude $m_t$ and directional grids $a_t$ each contribute 64 dimensions. The descriptor is
\begin{equation}
x_t=[h_t;g_t;e_t;m_t;a_t]\in\mathbb{R}^{768},
\end{equation}
then robustly standardized, projected to 32 dimensions using randomized PCA, and $\ell_2$-normalized.

ConsensusTAS divided the 20.9-s video into nine anonymous phases. In
Fig.~\ref{fig:demo}, the upper panel displays the predicted intervals,
while the lower panel shows the consensus boundary density averaged over seeds:
\begin{equation}
\bar d(t)=\frac{1}{3}\sum_{r\in\{7,17,27\}}d_r(t).
\end{equation}

The density peaks coincide with the decoded phase boundaries and provide a
visual indication of agreement across stochastic runs. To summarize the visual content of each phase, we extracted the medoid frame whose visual feature is closet to the phase's mean descriptor:

\begin{equation}
\mu_i
=
\frac{1}{|S_i|}
\sum_{t \in S_i} z_t ,
\end{equation}
\begin{equation}
t_i^{*}
=
\underset{t \in S_i}{\arg\min}
\left\lVert z_t-\mu_i \right\rVert_2 .
\end{equation}

\begin{table}[!t]
\caption{Tentative visual interpretation of the identified phases.}
\label{tab:phase_interpretation}
\centering\footnotesize
\setlength{\tabcolsep}{2pt}
\renewcommand{\arraystretch}{0.88}
\begin{tabular}{@{}ccp{0.62\columnwidth}@{}}
\toprule
Phase & Interval & Visual interpretation \\
\midrule
1 & 0.0--7.0 s   & Holding the brick, collecting and spreading mortar \\
2 & 7.0--9.3 s   & Completing mortar application \\
3 & 9.3--10.0 s  & Transition into placement \\
4 & 10.0--13.2 s & Inserting and initially positioning the closure brick \\
5 & 13.2--15.6 s & Pressing and aligning the brick \\
6 & 15.6--16.7 s & Short joint or edge adjustment \\
7 & 16.7--17.8 s & Continued alignment or mortar adjustment \\
8 & 17.8--18.6 s & Brief finishing movement \\
9 & 18.6--20.9 s & Final placement and hand withdrawal \\
\bottomrule
\end{tabular}
\end{table}

Figure~\ref{fig:bricklaying-medoids} presents the nine medoid frames in temporal
order. The inferred phases were evaluated post-hoc by a human reviewer and their interpretability and practical plausibility are shown in Table~\ref{tab:phase_interpretation}.

\section{Conclusions}

This study introduced \method{}, a label-free approach to TAS for long-horizon construction videos. Preliminary results demonstrated its superior performance on public benchmarks and produced plausible action segmentation of real-world construction activities. Future work will scale the method across diverse activities and establish a video benchmark for context-aware human–robot collaboration in long-horizon construction workflows.


\bibliographystyle{IEEEtran}
\bibliography{references}

\end{document}